\pdfoutput=1
\documentclass[11pt]{article}
\usepackage[final]{acl}
\usepackage{times}
\usepackage{latexsym}
\usepackage[T1]{fontenc}
\usepackage[utf8]{inputenc}
\usepackage{microtype}
\usepackage{inconsolata}
\usepackage{graphicx}
\usepackage[flushleft]{threeparttable}
\usepackage{booktabs} 
\usepackage{adjustbox}
\usepackage{array} 
\usepackage{multirow}
\usepackage{siunitx}
\usepackage[table]{xcolor}
\newcommand{\best}[1]{%
  \cellcolor{gray!20}\bfseries #1%
}
\newcolumntype{G}{@{}p{1.3em}@{}}
\newcolumntype{J}{@{}p{2.3em}@{}}
\usepackage{amsfonts}
\usepackage{subcaption}
\usepackage{caption}
\title{Bridging Discourse Treebanks with a Unified Rhetorical Structure Parser}
\author{Elena Chistova \\
  \texttt{chistova@isa.ru}
}
\begin{document}
\maketitle
\begin{abstract}
We introduce UniRST, the first unified RST-style discourse parser capable of handling 18 treebanks in 11 languages without modifying their relation inventories. To overcome inventory incompatibilities, we propose and evaluate two training strategies: \textit{Multi-Head}, which assigns separate relation classification layer per inventory, and \textit{Masked-Union}, which enables shared parameter training through selective label masking. We first benchmark mono-treebank parsing with a simple yet effective augmentation technique for low-resource settings. We then train a unified model and show that (1) the parameter efficient Masked-Union approach is also the strongest, and (2) UniRST outperforms 16 of 18 mono‑treebank baselines, demonstrating the advantages of a single-model, multilingual end-to-end discourse parsing across diverse resources.\footnote{Our models and code: \url{https://github.com/tchewik/UniRST}.}
\end{abstract}
\section{Introduction}
Rhetorical Structure Theory (RST) \cite{mann1987rhetorical} represents discourse as a hierarchical tree of elementary discourse units (EDUs) connected by rhetorical relations. Over the years, RST has inspired the creation of multiple discourse treebanks across different languages. However, large-scale annotated corpora are scarce and predominantly available in English. For other languages, the high cost of annotation and inconsistent guidelines have resulted in smaller, heterogeneous resources with incompatible relation inventories.
The English RST Discourse Treebank (RST-DT) \cite{carlson-etal-2001-building}, the primary benchmark for RST parsing, defines 56 fine-grained rhetorical relations, usually mapped to 18 coarse-grained classes for training and evaluation. Many discourse treebanks in other languages define considerably fewer relations. Aligning them with the RST-DT inventory often requires collapsing relations, such as merging \textsc{Cause} with \textsc{Effect}, \textsc{Contrast} with \textsc{Concession}, or \textsc{Elaboration} with \textsc{Entity-Elaboration}. This process erases distinctions that can be crucial for downstream applications such as coreference resolution, narrative analysis, and opinion mining. Moreover, when no direct equivalents exist, alignment is frequently based on surface-level label similarity, which compromises annotation reliability across languages.
End-to-end RST parsing involves three interconnected subtasks: EDU segmentation, tree structure prediction, and nuclearity and relation labeling. The definitions of these tasks are shaped by the relation inventory and constraints of each treebank. For instance, segmentation decisions can be influenced by fine-grained intra-sentential relations. Mono- or multinuclearity of certain overlapping relations (\textsc{Label\_NS}, \textsc{Label\_SN}, \textsc{Label\_NN}) varies across treebanks. When datasets with different inventories are merged and collapsed into a coarser label set, inconsistencies in relation definitions and nuclearity distributions can introduce substantial noise into both training and evaluation.
Despite these challenges, training on multiple treebanks offers clear benefits. RST-style parsers are known to generalize poorly across domains \cite{liu-zeldes-2023-cant}, and training a unified parsing model on all available treebanks may yield broader applicability. The skewed label distributions within individual corpora complicate model training, particularly in low-resource settings; pooling datasets with overlapping labels can mitigate this issue. Although larger treebanks provide sufficient data for accurate EDU segmentation and local relation labeling, they remain too small to support robust learning of global document structures. Leveraging all annotated structures across corpora can thus strengthen structural prediction. Altogether, these considerations motivate the development of universal discourse parsers that effectively integrate all available resources, regardless of language, genre, or domain.
In this work, we propose methods for building a unified RST parser from heterogeneous treebanks. Our contributions are:
\begin{enumerate}
    \item The first large-scale RST parsing study covering 18 treebanks in 11 languages.
    \item Data augmentation technique allowing for strong end-to-end mono-treebank RST parsing baselines even in low-resource settings.
    \item Two strategies for jointly modeling divergent relation inventories: Multi-Head and Masked-Union.
    \item Evaluations showing that: (i) dataset-specific segmentation heads are essential for handling varying EDU definitions; (ii) the Masked-Union approach enables sufficient model training by leveraging label overlap while respecting treebank-specific relation inventories, and (iii) our unified model outperforms 16 out of 18 mono-treebank baselines.
\end{enumerate}
\section{Related Work}
\paragraph{Cross-Lingual RST Parsing } Cross-lingual rhetorical structure parsing has gained increasing attention in recent years. \citet{braud-etal-2017-cross} introduced a unified set of coarse-grained (harmonized) rhetorical relations and presented the first data-driven cross-lingual RST parser, transferring across English, Brazilian Portuguese, Spanish, German, Basque, and Dutch. Their study demonstrated that rhetorical structure parsing from pre-segmented texts successfully transfers beyond English and across typologically diverse languages. Building on this foundation, \citet{liu-etal-2020-multilingual-neural} leveraged multilingual embeddings and proposed EDU-level machine translation to enrich training data. Subsequently, \citet{liu-etal-2021-dmrst} introduced DMRST, a unified framework performing joint EDU segmentation and discourse tree parsing, enabling end-to-end RST parsing evaluation across multiple languages under harmonized inventories. Extending this line of work, \citet{chistova-2024-bilingual} applied DMRST to parallel English–Russian data, highlighting the importance of aligned corpora for assessing cross-lingual transfer in the context of RST treebank incompatibilities.
\paragraph{Training on Incompatible Treebanks} Research on integrating incompatible treebanks has largely focused on syntax parsing. Early work by
\citet{johansson-2013-training} introduced two adaptation techniques for training syntax parsers on treebanks with differing annotation schemes. Their methods involved concatenating the feature spaces of two treebanks and using a parser trained on one treebank to guide the other. These approaches were applied to treebanks pairs within the same language (German, Swedish, Italian, and English). \citet{stymne-etal-2018-parser} explored three strategies: treebank concatenation with and without fine-tuning, and the inclusion of treebank-specific embeddings. Their results showed consistent improvements in dependency parsing for most of the nine languages evaluated when using treebank-specific embeddings. A similar approach was applied by \citet{barry-etal-2019-cross} to train a cross-lingual parser for low-resource Faroese syntax parsing. \citet{johansson-adesam-2020-training} trained a Swedish constituency parser on six incompatible treebanks by sharing word representations across corpora while maintaining separate neural parsing modules for each treebank, thus accommodating both constituency and dependency annotations. \citet{kankanampati-etal-2020-multitask} leveraged two Arabic dependency treebanks to build a parser with a unified attachment scorer. \citet{sayyed-dakota-2021-annotations} conducted multilingual experiments with treebank-specific biaffine parsing layers for UD and SUD syntactic annotations, ultimately finding that combining distinct annotation schemes could degrade parsing performance.
Notably, in syntactic parsing, terminal nodes correspond to words, so efforts to resolve annotation inconsistencies are confined to structure building and label assignment. In contrast, rhetorical structure parsing additionally requires segmentation, which is affected by treebank-specific constraints on elementary discourse units. In our work, we aim to develop the first end-to-end RST parser benefiting from each annotation scheme in a wide range of diverse discourse treebanks. 
\section{UniRST}
\begin{figure*}[t!]
  \centering
  \begin{subfigure}[t]{0.3\textwidth}
    \centering
    \adjustbox{valign=t}{\includegraphics[height=6cm]{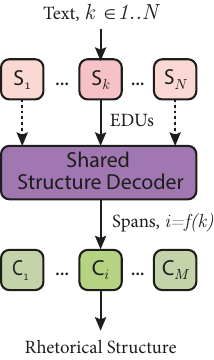}}
    \caption{\textsc{MH} architecture}
    \label{fig:mh}
  \end{subfigure}
  \hfill
  \begin{subfigure}[t]{0.6\textwidth}
    \centering
    \adjustbox{valign=t}{\includegraphics[height=6cm]{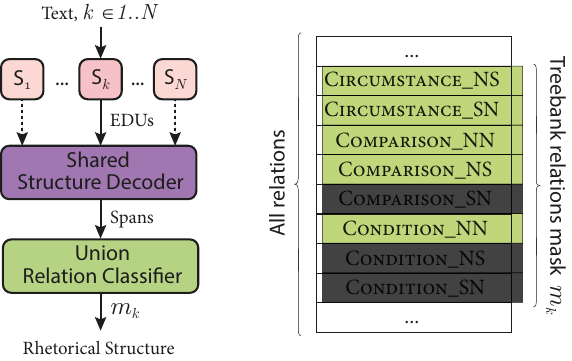}}
    \caption{\textsc{MU} architecture}
    \label{fig:mu}
  \end{subfigure}
  \caption{Model variants in the UniRST framework. (a) Multi-Head: independent classifiers per relation inventory. (b) Masked-Union: shared classifier with treebank-specific label masking.}
  \label{fig:architecture}
\end{figure*}
We address joint training over heterogeneous RST corpora while preserving each treebank’s native relation inventory, EDU segmentation, and relational definitions. Building on the DMRST architecture \cite{liu-etal-2021-dmrst}, we explore extensions that enable training across incompatible treebanks. Specifically, we propose two strategies: \textit{Multi-Head} (\textsc{MH}), which maintains separate classification heads per inventory, and \textit{Masked-Union} (\textsc{MU}), which uses a single classifier constrained by treebank-specific masks. For reference, we additionally implement \textit{Unmasked-Union} (\textsc{UU}), which lacks label masking and serves as a lower bound. Unless otherwise noted, models use treebank-specific segmentation heads, though shared segmentation is also tested. Figure~\ref{fig:architecture} illustrates the architectures.
\subsection{DMRST}
DMRST \cite{liu-etal-2021-dmrst} is an end-to-end RST parsing model that integrates EDU segmentation, discourse tree construction, and relation/nuclearity labeling. Its pipeline has four stages: (1) a pretrained language model encodes input tokens, (2) an LSTM-CRF module detects EDU boundaries, (3) a recurrent pointer network decoder constructs the discourse tree, and (4) a biaffine classifier assigns nuclearity and relation labels. The model is trained jointly, with dynamically weighted loss balancing segmentation, structure prediction, and labeling. This unified design enables consistent end-to-end parsing. 
UniRST extends this backbone to multi-treebank training. The pretrained encoder and recurrent decoder are shared across corpora, while segmentation and relation classification are treebank-tailored. This design aims to achieve robust structural prediction while respecting each corpus’s definitions and constraints.
\subsection{Multi-Head (\textsc{MH})} 
Our first method for multi-inventory RST parsing assigns a separate classification head to each distinct relation inventory. Given the set of inventories $\mathcal{G}=\{G_1,\dots,G_M\}$, treebanks sharing the same inventory (e.g., \texttt{eng.gum}, \texttt{rus.rrg}, \texttt{zho.gcdt}) share a relation/nuclearity classifier $\mathbf{W}^{(m)}\in\mathbb{R}^{d\times|G_m|}$. In this configuration, cross-treebank information about relation and nuclearity is exchanged only implicitly, through fine-tuning of the language model and shared structural decoder.
\subsection{Masked-Union  (\textsc{MU})} 
Let $\mathcal{U}=\bigcup_{k}\mathcal{L}_{T_k}$ be the unified set of all relation types across treebanks. \textsc{MU} employs a single shared classifier $\mathbf{W}\in\mathbb{R}^{d\times|\mathcal{U}|}$ that predicts over this unified label space. To enforce inventory constraints, for each treebank $T_k$, we apply a binary mask $\mathbf{m}^{(k)} \in \{ -1 \times 10^{9}, 1 \}^{|\mathcal{U}|}$ to the classifier logits. This parameter-efficient design promotes explicit parameter sharing and enables direct transfer for overlapping relations (e.g., \textsc{Elaboration\_NS}) across all components of the model.
\subsection{Unmasked-Union (\textsc{UU})} 
\textsc{UU} mirrors the \textsc{MU} architecture but omits the treebank-specific masking, thereby allowing predictions over the entire concatenated label set without restriction. Consequently, it can produce labels that do not exist in the target corpus, limiting its practical utility. We include \textsc{UU} scores as a lower-bound baseline.
\section{Data}  
\begin{table*}[ht]
\centering
\small
\begin{adjustbox}{max width=\textwidth}
\begin{tabular}{lccccccc}
\toprule
\textbf{Treebank} & \textbf{Language}  & \textbf{\# Docs} & \textbf{\# Tokens} & \textbf{\# EDUs} & \textbf{\# Labels} & \textbf{\# Classes} & \textbf{\# Rels} \\
\midrule
ces.crdt      \citeyearpar{polakova-etal-2024-developing}        & Czech   & 54   & 14{,}623       & 1{,}345   & 23  & 34  & 1{,}288 \\
deu.pcc       \citeyearpar{stede-neumann-2014-potsdam}           & German  & 176  & 32{,}836       & 2{,}842   & 25  & 37  & 2{,}665 \\
eng.gentle    \citeyearpar{aoyama-etal-2023-gentle}  (test only) & English & 26   & 17{,}799       & 2{,}328   & 15  & 27  & 2{,}552 \\
eng.gum v11.1 \citeyearpar{zeldes-etal-2025-erst}                & English & 255  & 250{,}290      & 34{,}428  & 15  & 27  & 32{,}173 \\
eng.oll       \citeyearpar{potter2008interactional}              & English & 327  & 46{,}177       & 3{,}026   & 21  & 35  & 2{,}716 \\
eng.rstdt     \citeyearpar{RSTDT-LDC}                            & English & 385  & 205{,}829      & 21{,}789  & 18  & 42  & 21{,}404 \\
eng.sts       \citeyearpar{potter2008interactional}              & English & 150  & 70{,}422       & 3{,}208   & 21  & 35  & 3{,}058 \\
eng.umuc      \citeyearpar{zaczynska-stede-2024-rhetorical}      & English & 87   & 61{,}292       & 5{,}421   & 28  & 46  & 5{,}334 \\
eus.ert       \citeyearpar{IruskietaAranzabeIlarrazaEtAl2013}    & Basque  & 88   & 45{,}780       & 2{,}509   & 24  & 31  & 2{,}421 \\
fas.prstc     \citeyearpar{shahmohammadi2021persian}             & Persian & 150  & 66{,}694       & 5{,}789   & 18  & 26  & 5{,}638 \\
fra.annodis   \citeyearpar{afantenos-etal-2012-empirical}        & French  & 86   & 32{,}699       & 3{,}307   & 18  & 20  & 3{,}221 \\
nld.nldt      \citeyearpar{redeker-etal-2012-multi}              & Dutch  & 80   & 24{,}898       & 2{,}326   & 27  & 45  & 2{,}246 \\
por.cstn      \citeyearpar{CardosoMazieroRosarioCastroJorgeEtAl2011} & Portuguese  & 140 & 58{,}793 & 5{,}527   & 22  & 38  & 5{,}387 \\
rus.rrg       \citeyearpar{chistova-2024-bilingual}              & Russian  & 213  & 172{,}405      & 25{,}222  & 15  & 27  & 25{,}010 \\
rus.rrt       \citeyearpar{PisarevskayaEtAl2017}                 & Russian  & 233  & 262{,}495      & 28{,}247  & 17  & 25  & 25{,}892 \\
spa.rststb    \citeyearpar{da-cunha-etal-2011-development}       & Spanish  & 267  & 58{,}717       & 3{,}351   & 29  & 43  & 3{,}084 \\
spa.sctb      \citeyearpar{cao-etal-2018-rst}                    & Spanish  & 50   & 16{,}515       & 744       & 20  & 26  & 694 \\
zho.gcdt      \citeyearpar{peng-etal-2022-gcdt}                  & Chinese  & 50   & 62{,}905       & 9{,}403   & 15  & 28  & 9{,}345 \\
zho.sctb      \citeyearpar{cao-etal-2018-rst}                    & Chinese  & 50   & 15{,}496       & 744       & 20  & 26  & 684 \\
\bottomrule
\end{tabular}
\end{adjustbox}
\caption{Treebank statistics.}
\label{tab:treebanks}
\end{table*}
This study leverages training data from 18 RST treebanks covering 11 languages, aiming to create the most universal end-to-end RST parser to date. The treebanks span Czech, German, English, Basque, Persian, French, Dutch, Brazilian Portuguese, Russian, Spanish, and Chinese.  
Treebank statistics are summarized in Table~\ref{tab:treebanks}. 
For the English RST-DT benchmark, we adopt the coarse-grained relation labels used in prior work.  Corpora annotated using the GUM RST schema (\texttt{eng.gum}, \texttt{zho.gcdt}, \texttt{rus.rrg}) retain their predefined coarse-grained labels. For other corpora, if applicable, we merged infrequent classes (less than 10 instances) with related ones based on nuclearity, following the mapping suggested by \citet{braud-etal-2017-cross}. This ensures both label diversity and sufficient representation for training. Detailed class distributions are illustrated in  Appendix~\ref{sec:appendix_labels}.  
To ensure consistency and reproducibility, we use the standardized training, validation, and test splits\footnote{We employ the open version of \texttt{eus.ert} treebank from \url{https://ixa2.si.ehu.eus/diskurtsoa/en/}, containing 88 annotations.} provided by the DISRPT 2025 shared task for segmentation, connective identification, and relation classification across discourse annotation frameworks.
\subsection{Data Augmentation}
While several large RST treebanks dominate end-to-end discourse parsing research, smaller corpora remain underutilized due to limited training data. To address this gap and establish strong mono-treebank baselines, we propose a simple yet effective data augmentation technique to improve performance in low-resource settings. Crucially, our method enriches training data without modifying the original texts or local annotations. 
DMRST model employs a recurrent structure prediction module that relies heavily on contextual signals. As each annotated tree yields a single training instance, the number of examples is limited, particularly in smaller treebanks. To address this, we introduce an augmentation approach based on extracting structurally coherent subtrees from annotated documents. While these subtrees omit full-document context, their internal discourse structure remains valid and informative.
Our procedure involves: (1) identifying sentence boundaries to avoid extracting subtrees spanning sentence fragments; (2) extracting all connected subtrees not spanning sentence fragments and including at least three rhetorical relations; and (3) sampling a proportion $p_{\text{aug}}$ of these subtrees for augmentation. Sampling is critical to prevent overfitting, particularly for the segmentation subtask.
This augmentation allows the model to train on a wider range of partial structures, potentially improving end-to-end RST parsing training in low-resource settings. We set $p_{\text{aug}}$ to 50\% to enrich the training data multifold.\footnote{For RST-DT, $p_{\text{aug}} = 50\%$ produces 5.4 times more training samples. Over all treebanks, it multiplies number of training samples by 7.7.}
\begin{table*}[ht]
  \centering
  \small
  \setlength{\tabcolsep}{2.7pt}
  \begin{adjustbox}{max width=\textwidth}
  \begin{tabular}{
  l                  
  *{9}{S[table-format=2.1]} 
  G
  *{9}{S[table-format=2.1]} 
}
  \toprule
  & \multicolumn{9}{c}{\bfseries Baseline}
    & 
    & \multicolumn{9}{c}{\bfseries + Augmentation} \\
  \cmidrule(lr){2-10}  \cmidrule(lr){12-20}
  & \multicolumn{4}{c}{\bfseries Gold seg}
    & \multicolumn{5}{c}{\bfseries End‐to‐end}
    &
    & \multicolumn{4}{c}{\bfseries Gold seg}
    & \multicolumn{5}{c}{\bfseries End‐to‐end} \\
    \textbf{Treebank}
      & \textbf{S} & \textbf{N} & \textbf{R} & \textbf{Full}
      & \textbf{Seg} & \textbf{S} & \textbf{N} & \textbf{R} & \textbf{Full}
      &
      & \textbf{S} & \textbf{N} & \textbf{R} & \textbf{Full}
      & \textbf{Seg} & \textbf{S} & \textbf{N} & \textbf{R} & \textbf{Full} \\
    \midrule
    ces.crdt            & \best{60.2}  & \best{31.1}  & \best{18.2}  & 16.9  
                & 90.1  & \best{47.3}  & \best{24.0}  & \best{13.7}  & \best{12.3}
                &       & 58.9  & \best{31.1}  & 18.0  & \best{17.1}  
                & \best{90.6}  & 46.2  & 23.4  & 11.5  & 11.2  \\
    deu.pcc             & \best{68.7}  & 38.8  & 24.7  & 23.6  
                & 95.2  & 58.9  & 33.9  & 21.0  & 20.0
                &       & 67.2  & \best{42.7}  & \best{26.4}  & \best{25.6}  
                & \best{96.0}  & \best{59.7}  & \best{36.9}  & \best{21.7}  & \best{21.2}  \\
    eng.gum             & \best{73.3}  & \best{60.5}  & \best{52.6}  & \best{51.5}  
                & \best{95.5}  & \best{66.9}  & \best{55.4}  & \best{48.3}  & \best{47.4}
                &       & 72.8  & 59.9  & \best{52.6}  & 51.4  
                & 95.2  & 66.1  & 54.3  & 47.9  & 46.9  \\
    eng.oll             & \best{65.9}  & \best{48.2}  & \best{29.5}  & \best{29.3}  
                & 89.7  & 51.4  & \best{36.7}  & 21.9  & 21.5
                &       & 61.8  & 43.2  & 29.0  & 28.4  
                & \best{91.2}  & \best{54.1}  & 36.4  & \best{24.5}  & \best{24.0}  \\
    eng.rstdt           & 77.5  & 66.6  & 56.1  & 54.6  
                & 97.6  & 73.8  & 63.4  & 53.3  & 51.8 
                &       & \best{78.3}  & \best{67.5}  & \best{57.0}  & \best{55.2}  
                & \best{97.7}  & \best{74.9}  & \best{64.5}  & \best{54.6}  & \best{52.9}  \\
    eng.sts             & \best{46.5}  & \best{35.1}  & \best{21.7}  & \best{21.1}  
                & \best{89.7}  & \best{38.2}  & \best{28.8}  & \best{18.2}  & \best{18.0}  
                &       & 44.1  & 32.9  & 20.5  & 19.6  & 88.4  & 33.5  & 24.5  & 16.1  & 15.7  \\
    eng.umuc            & \best{68.8}  & \best{50.8}  & \best{33.1}  & \best{32.4}  
                & \best{89.6}  & \best{51.2}  & \best{36.9}  & \best{24.3}  & 23.7  
                &       & 67.1  & 48.4  & 31.8  & 31.2  
                & 89.0  & 49.1  & 35.1  & \best{24.3}  & \best{24.0}  \\
    eus.ert             & \best{71.0}  & \best{47.3}  & \best{29.9}  & \best{29.2}  
                & \best{89.7}  & \best{54.8}  & \best{38.0}  & \best{23.1}  & \best{22.7}  
                &       & 66.5  & 44.5  & 25.7  & 25.7  
                & 89.0  & 52.3  & 35.3  & 19.9  & 19.8  \\
    fas.prstc           & 65.0  & \best{51.3}  & 40.2  & 40.1  
                & \best{93.8}  & \best{55.3}  & \best{44.6}  & \best{34.4}  & \best{34.4} 
                &       & \best{65.3}  & 50.9  & \best{40.7}  & \best{40.4}  
                & \best{93.8}  & \best{55.3}  & 42.9  & 34.3  & 34.0  \\
    fra.annodis         & \best{62.5}  & \best{51.6}  & \best{33.0}  & \best{33.0}  
                & \best{92.1}  & \best{53.2}  & \best{44.7}  & \best{28.6}  & \best{28.6}  
                &       & \best{62.5}  & 51.4  & 32.9  & 32.9  
                & 91.5  & 52.4  & 43.3  & 27.6  & 27.6  \\
    nld.nldt            & \best{63.8}  & \best{47.4}  & \best{30.7}  & \best{29.1}
                & 96.3  & \best{58.2}  & \best{42.9} & 28.5  & 26.9  
                &       & 61.7  & 46.4  & 30.6  & 28.8 
                & \best{96.4}   & 57.2  & 42.8  & \best{28.9}  & \best{27.3}  \\
    por.cstn            & 76.0  & 62.2  & 50.9  & 50.8  
                & 93.9  & \best{68.2}  & \best{53.3}  & \best{43.9}  & \best{43.8}
                &       & \best{76.1}  & 61.6   & 49.9  & 49.9  
                & \best{94.0}  & 66.3  & \best{53.3}   & 43.0  & 43.0  \\
    rus.rrg             & \best{71.2}  & \best{57.2}  & \best{49.4}  & \best{48.2}  
                & \best{97.2}  & \best{67.6}  & \best{54.3}  & \best{47.1}  & \best{46.0}  
                &       & 70.3  & 55.9  & 47.9  & 46.8  
                & 96.8  & 65.6  & 52.2  & 44.9  & 43.9  \\
    rus.rrt             & 79.7  & 61.4  & 51.5  & 51.3  
                & \best{91.1}  & 62.8  & 49.0  & 41.4  & 41.3  
                &       & \best{80.0}  & \best{61.9}  & \best{52.2}  & \best{51.9}  
                & 91.0  & \best{63.0}  & \best{49.9}  & \best{42.5}  & \best{42.3}  \\
    spa.rststb          & 68.1  & 52.2  & 35.0  & 35.0  
                & 91.5  & 54.9  & 40.9  & 28.1  & 28.1  
                &       & \best{71.1}  & \best{54.4}  & \best{38.9}  & \best{38.9}  
                & \best{91.9}  & \best{57.7}  & \best{44.0}  & \best{32.8}  & \best{32.8}  \\
    spa.sctb            & \best{66.7}  & 41.5  & \best{35.2}  & \best{35.2}  
                & 74.1  & 34.3  & 25.4  & 23.1  & 23.1  
                &       & \best{66.7}  & \best{43.6}  & 31.7  & 31.7  
                & \best{84.1}  & \best{47.5}  & \best{35.3}  & \best{27.3}  & \best{27.3}  \\
    zho.gcdt            & \best{76.3}  & 58.1  & \best{52.3}  & \best{50.7}  
                & 91.2  & 61.0  & 46.1  & 40.9  & 39.6  
                &       & 75.3  & \best{58.2}   & 51.9  & 50.4  
                & \best{91.9}  & \best{64.0}  & \best{49.5}  & \best{43.5}  & \best{42.3}  \\
    zho.sctb            & \best{66.7}  & 37.7  & 32.1  & 32.1  
                & 91.1  & \best{56.3}  & 34.3  & 28.5  & 28.5 
                &       & 60.2  & \best{40.9}  & \best{32.3}  & \best{32.3}  
                & \best{92.5}  & 52.3  & \best{38.1}  & \best{29.6}  & \best{29.6}  \\
    \bottomrule
  \end{tabular}
  \end{adjustbox}
  \caption{Performance of the treebank-specific models, with and without train data augmentation.}
  \label{tab:mono-results}
\end{table*}
\section{Experimental Setup}
We employ \texttt{xlm-roberta-large} as the multilingual encoder across all experiments. The batch size is set to 2, with a hidden size of 200 for the segmenter and 512 for the parsing module. The DMRST model is trained with a learning rate of 1e-5, while the encoder is fine-tuned using a learning rate of 2e-5.  Early stopping is set to a patience of 5 in mono-treebank settings and reduced to 3 in UniRST due to the larger concatenated dataset.
Evaluation follows the original Parseval metrics for rhetorical structure parsing, with micro F1 scores reported for segmentation (Seg), span (S), relation (R), nuclearity (N), and full structure (Full). Each model is trained using three different random seeds, and all reported results are averaged across these runs.
\section{Experimental Results}
\subsection{Mono-Treebank Evaluations}
\label{ssec:monotreebank}
Table~\ref{tab:mono-results} reports the performance of treebank-specific models trained with and without data augmentation. Augmentation yielded substantial gains on smaller corpora such as \texttt{eng.oll}, \texttt{spa.sctb}, \texttt{zho.sctb}, and \texttt{zho.gcdt}, but improvements were not uniform across all treebanks. Interestingly, on the \texttt{eng.rstdt} benchmark with diverse document lengths, augmentation led to an average 1.1\% F1 improvement in unlabeled structure prediction (S), highlighting its potential even for larger datasets. On the other hand, the data augmentation resulted in performance degradation on two large-scale GUM-based corpora (\texttt{eng.gum}, \texttt{rus.rrg}), likely due to segmenter overfitting on long documents. Overall, augmentation yielded the best mono-treebank parsing performance on 10 of the 18 treebanks.
For comparison, Appendix~\ref{sec:reference-results} summarizes previous end-to-end RST parsing results on eight treebanks. DMRST+ denotes the architecture used as a baseline in this work. 
\begin{table}[t]
\setlength{\tabcolsep}{4pt}
\centering
\small
\begin{tabular}{l ccc G ccc}
\toprule
\multirow{2}{*}{\textbf{Model}} & \multicolumn{3}{c}{\textbf{In-treebank}} & & \multicolumn{3}{c}{\textbf{All avg.}} \\
 & \textbf{Seg} & \textbf{S} & \textbf{N} & & \textbf{Seg} & \textbf{S} & \textbf{N} \\
\midrule
ces.crdt      & 90.5 & 49.5 & 24.5 &  & 76.4 & 32.6 & 16.0 \\
deu.pcc       & 96.0 & 59.7 & 36.9 &  & 76.6 & 32.9 & 19.3 \\
eng.gum       & 95.5 & 66.9 & 55.4 &  & 78.9 & 41.3 & 30.2 \\
eng.oll       & 91.2 & 54.1 & 35.4 &  & 77.3 & 31.9 & 18.7 \\
eng.rstdt     & 97.7 & 74.9 & 64.5 &  & 78.4 & 40.2 & 28.8 \\
eng.sts       & 89.7 & 38.2 & 32.6 &  & 75.8 & 29.1 & 16.8 \\
eng.umuc      & 89.6 & 51.2 & 36.9 &  & 77.6 & 35.1 & 22.4 \\
eus.ert       & 89.7 & 54.8 & 38.0 &  & 77.5 & 33.8 & 18.4 \\
fas.prstc     & 93.8 & 58.3 & 44.6 &  & 78.3 & 36.2 & 20.3 \\
fra.annodis   & 92.1 & 53.2 & 44.7 &  & 77.5 & 32.6 & 16.5 \\
nld.nldt      & 96.4 & 57.0 & 43.9 &  & 80.2 & 35.9 & 22.0 \\
por.cstn      & 94.1 & 68.8 & 57.2 &  & 78.2 & 37.1 & 25.2 \\
rus.rrg       & 97.2 & 67.6 & 54.3 &  & 80.0 & 40.9 & 28.7 \\
rus.rrt       & 91.0 & 63.0 & 49.9 &  & 81.2 & 40.9 & 27.3 \\
spa.rststb    & 91.9 & 57.7 & 44.0 &  & 78.5 & 35.4 & 22.6 \\
spa.sctb      & 84.1 & 47.5 & 35.3 &  & 74.0 & 29.9 & 16.2 \\
zho.gcdt      & 91.9 & 64.0 & 49.5 &  & 76.9 & 36.7 & 23.3 \\
zho.sctb      & 92.5 & 52.3 & 38.1 &  & 57.4 & 17.1 & 10.2 \\
\midrule
UniRST        & ---  & ---  & ---  &  & 92.9  & 60.7  & 47.3 \\
\bottomrule
\end{tabular}
\caption{Evaluation across all treebanks. We only assess segmentation (Seg), unlabeled structure construction (S), and nuclearity assignment (N), as relation inventories are incompatible.}
\label{tab:mono-transfer}
\end{table}
\begin{table*}[ht]
  \centering
  \small
  \begin{adjustbox}{max width=\textwidth}
    \begin{tabular}{l c cccc J ccccc}
    \toprule
    \multirow{2}{*}{\textbf{Method}}  & \multirow{2}{*}{\textbf{Segmentation}}
     & \multicolumn{4}{c}{\textbf{Gold seg}} & & \multicolumn{5}{c}{\textbf{End-to-end}} \\
                   &  & \textbf{S}  & \textbf{N}  & \textbf{R}  & \textbf{Full}  & & \textbf{Seg}  & \textbf{S}  & \textbf{N}  & \textbf{R}  & \textbf{Full}  \\
    \midrule
    \textsc{MH}    & single       & 73.5  & 58.4   & 47.6  & 46.6   & & 93.4  & 63.4   & 50.7   & 41.7   & 40.8   \\
                   & multiple     & 73.6  & 59.0   & 48.5  & 47.6   & & 93.7  & 63.7   & 51.3   & 42.4   & 41.6   \\ \midrule
    \textsc{UU}    & single       & 73.8  & 58.7   & 47.0  & 46.8   & & 93.4  & 64.3   & 51.9   & 42.8   & 41.8   \\ \midrule
    \textsc{MU}    & single       & 74.1  & 59.3   & 48.8  & 47.8   & & 93.7  & 64.5   & \textbf{52.1}   & 43.2   & 42.3   \\ 
                   & multiple     & \textbf{74.4}  & \textbf{59.6}   & \textbf{49.3}  & \textbf{48.3}   & & \textbf{93.9}  & \textbf{64.8}   & \textbf{52.1}   & \textbf{43.4}   & \textbf{42.5}   \\
    \bottomrule
    \end{tabular}
  \end{adjustbox}
  \caption{Performance of the UniRST model in different setups.}
  \label{tab:unirst-main}
\end{table*}
\begin{table*}[t]
\centering
\small
\setlength{\tabcolsep}{6pt}
\renewcommand{\arraystretch}{1.15}
\begin{tabular}{llllll}
\toprule
\textbf{Treebank} & \textbf{Seg} & \textbf{S} & \textbf{N} & \textbf{R} & \textbf{Full} \\
\midrule
ces.crdt    & 94.2 \textcolor{gray}{ (+4.1)} & 57.9 \textcolor{gray}{ (+10.6)} & 38.6 \textcolor{gray}{ (+14.6)} & 27.3 \textcolor{gray}{ (+3.3)}  & 26.8 \textcolor{gray}{ (+14.5)} \\
deu.pcc     & 96.5 \textcolor{gray}{ (+0.5)} & 66.3 \textcolor{gray}{ (+6.6)}  & 45.5 \textcolor{gray}{ (+8.6)}  & 32.8 \textcolor{gray}{ (+11.1)} & 31.1 \textcolor{gray}{ (+9.9)} \\
eng.gum     & 95.2 \textcolor{gray}{ (-0.3)} & 66.7 \textcolor{gray}{ (-0.2))} & 54.7 \textcolor{gray}{ (-0.7)}  & 48.0 \textcolor{gray}{ (-0.3)}  & 46.9 \textcolor{gray}{ (-0.5)} \\
eng.oll     & 93.8 \textcolor{gray}{ (+2.6)} & 56.7 \textcolor{gray}{ (+2.6)}  & 40.6 \textcolor{gray}{ (+4.2)}  & 27.6 \textcolor{gray}{ (+3.1)}  & 27.1 \textcolor{gray}{ (+3.1)} \\
eng.rstdt   & 97.8 \textcolor{gray}{ (+0.1)} & 75.6 \textcolor{gray}{ (+0.7)}  & 65.1 \textcolor{gray}{ (+0.6)}  & 55.2 \textcolor{gray}{ (+0.6)}  & 53.5 \textcolor{gray}{ (+0.6)} \\
eng.sts     & 91.0 \textcolor{gray}{ (+1.3)} & 40.4 \textcolor{gray}{ (+2.2)}  & 30.7 \textcolor{gray}{ (+1.9)}  & 19.4 \textcolor{gray}{ (+1.2)}  & 18.8 \textcolor{gray}{ (+0.8)} \\
eng.umuc    & 88.8 \textcolor{gray}{ (-0.8)} & 52.0 \textcolor{gray}{ (+0.8)}  & 40.1 \textcolor{gray}{ (+3.2)}  & 26.1 \textcolor{gray}{ (+1.8)}  & 25.6 \textcolor{gray}{ (+1.9)} \\
eus.ert     & 92.0 \textcolor{gray}{ (+2.3)} & 62.8 \textcolor{gray}{ (+8.0)}  & 47.4 \textcolor{gray}{ (+9.4)}  & 35.4 \textcolor{gray}{ (+12.3)} & 35.3 \textcolor{gray}{ (+12.6)} \\
fas.prstc   & 94.6 \textcolor{gray}{ (+0.8)} & 61.7 \textcolor{gray}{ (+6.4)}  & 50.2 \textcolor{gray}{ (+5.6)}  & 40.7 \textcolor{gray}{ (+6.3)}  & 40.5 \textcolor{gray}{ (+6.1)} \\
fra.annodis  & 90.9 \textcolor{gray}{ (-1.2)} & 58.1 \textcolor{gray}{ (+4.9)}  & 47.3 \textcolor{gray}{ (+2.6)}  & 31.1 \textcolor{gray}{ (+2.5)}  & 30.7 \textcolor{gray}{ (+2.1)} \\
nld.nldt    & 97.6 \textcolor{gray}{ (+1.2)} & 59.3 \textcolor{gray}{ (+2.1)}  & 45.3 \textcolor{gray}{ (+2.5)}  & 33.5 \textcolor{gray}{ (+4.6)}  & 31.7 \textcolor{gray}{ (+4.4)} \\
por.cstn    & 94.3 \textcolor{gray}{ (+0.4)} & 67.7 \textcolor{gray}{ (-0.5)}  & 54.9 \textcolor{gray}{ (+1.6)}  & 45.7 \textcolor{gray}{ (+1.8)}  & 45.4 \textcolor{gray}{ (+1.6)} \\
rus.rrg     & 96.5 \textcolor{gray}{ (-0.7)} & 66.8 \textcolor{gray}{ (-0.8)}  & 53.5 \textcolor{gray}{ (-0.8)}  & 45.5 \textcolor{gray}{ (-1.6)}  & 44.1 \textcolor{gray}{ (-1.9)} \\
rus.rrt     & 90.6 \textcolor{gray}{ (-0.4)} & 63.0 \textcolor{gray}{ (0.0)}   & 49.8 \textcolor{gray}{ (-0.1)}  & 42.6 \textcolor{gray}{ (+0.1)}  & 42.4 \textcolor{gray}{ (+0.1)} \\
spa.rststb  & 92.5 \textcolor{gray}{ (+0.6)} & 63.5 \textcolor{gray}{ (+5.8)}  & 50.1 \textcolor{gray}{ (+6.1)}  & 35.3 \textcolor{gray}{ (+2.5)}  & 35.2 \textcolor{gray}{ (+2.4)} \\
spa.sctb    & 86.0 \textcolor{gray}{ (+1.9)} & 55.8 \textcolor{gray}{ (+8.3)}  & 48.0 \textcolor{gray}{ (+12.7)} & 40.8 \textcolor{gray}{ (+13.5)} & 40.8 \textcolor{gray}{ (+13.5)} \\
zho.gcdt    & 92.1 \textcolor{gray}{ (+0.2)} & 62.9 \textcolor{gray}{ (-1.1)}  & 48.7 \textcolor{gray}{ (-0.8)}  & 44.0 \textcolor{gray}{ (+0.5)}  & 42.7 \textcolor{gray}{ (+0.4)} \\
zho.sctb    & 94.3 \textcolor{gray}{ (+1.8)} & 64.3 \textcolor{gray}{ (+12.0)} & 50.5 \textcolor{gray}{ (+12.4)} & 40.7 \textcolor{gray}{ (+11.1)} & 40.7 \textcolor{gray}{ (+11.1)} \\
\bottomrule
\end{tabular}
\caption{
UniRST performance per treebank. Improvements over the strongest mono-treebank baseline, as listed in Table~\ref{tab:mono-results}, are shown in parentheses.
}
\label{tab:unirst-by-treebank}
\end{table*}
To assess generalization, each best-performing treebank-specific model was evaluated on all 18 corpora. Table~\ref{tab:mono-transfer} reveals a consistent transferability gap: models tend to overfit to treebank-specific language, domains, relation usage, and document styles. Segmentation scores also decline in transfer settings, though less severely than Span or Nuclearity scores. In certain cases, however (e.g., \texttt{eng.oll}, \texttt{eng.gum}), segmentation drops sharply, reflecting variation in EDU definitions across corpora. Despite strong in-treebank Span F1 (e.g., 74.9\% for \texttt{eng.rstdt}, 68.8\% for \texttt{por.cstn}), transfer performance degrades substantially (dropping to 40.2\% and 37.1\%, respectively). This disparity demonstrates that in-domain success is a poor indicator of cross-corpus robustness and highlights the need for more generalizable RST parsers, such as UniRST.
\subsection{UniRST}
Performance of the Multi-Head and Masked-Union strategies is reported in Table~\ref{tab:unirst-main}. UniRST performs best when segmentation is handled by treebank-specific heads, which capture differences in EDU annotation schemes, whereas a universal segmentation head primarily learns broader segmentation patterns. The Masked-Union (\textsc{MU}) strategy consistently outperforms Multi-Head (\textsc{MH}), offering both greater efficiency and higher parsing accuracy. Its masking mechanism ensures that each treebank’s inventory is respected, while still enabling transfer for overlapping relations, which in turn improves parsing performance over the unmasked baseline. The strongest configuration is \textsc{MU} with treebank-specific segmentation heads. We refer to this variant as “UniRST” throughout the remainder of the paper.
As shown in Table~\ref{tab:mono-transfer}, UniRST achieves higher average performance across combined test set compared to any mono-treebank parser. This demonstrates the robustness of UniRST model as a cross-lingual parser capable of learning shared representations that generalize effectively across diverse RST corpora.
Detailed results by treebank are provided in Table~\ref{tab:unirst-by-treebank}. The unified model outperforms the strongest mono-treebank baselines on 16 out of 18 treebanks. Notable improvements in end-to-end Full F1 are observed across most datasets, particularly for smaller-scale treebanks such as \texttt{ces.crdt}, \texttt{deu.pcc}, \texttt{eus.ert}, \texttt{spa.sctb}, \texttt{zho.gcdt}, and \texttt{zho.sctb}. Similar to data augmentation in mono-treebank training, joint training does not benefit the large-scale \texttt{eng.gum} and \texttt{rus.rrg} corpora, whose annotations appear sufficient on their own. Importantly, the performance drop on \texttt{eng.gum} under joint training remains marginal.
The only corpus where UniRST fails to exceed 50\% Span F1 and 25\% Full F1 is \texttt{eng.sts}. Given the limited documentation of this dataset, the cause is unclear, but the low scores may stem from poor inter-annotator agreement or inconsistently applied segmentation and structural constraints. Joint training nonetheless improved performance, suggesting that it provides some stabilization even under noisy conditions. Across nine corpora in English, Persian, Portuguese, Russian, Spanish, and Chinese, UniRST achieves more than 40\% Full end-to-end F1 while preserving original relation inventories. 
To further assess out-of-domain generalization, we evaluate GUM-compatible models on the GENTLE benchmark, which follows GUM annotation guidelines.\footnote{GENTLE includes annotations for eight unconventional genres: dictionary entries, esports commentaries, legal documents, medical notes, poetry, mathematical proofs, syllabuses, and threat letters. None of these genres are represented in the training corpora used in this work.} As shown in Table~\ref{tab:gentle}, UniRST achieves the highest Full end-to-end parsing score. The \texttt{eng.gum} model performs best in segmentation (93.0\% F1) and structure prediction (58.0\%) due to its alignment with GENTLE’s language and annotation conventions. However, UniRST outperforms it on Relation and Full F1, highlighting the benefits of shared relation classification training across multiple treebanks. Notably, UniRST supports 11 languages, while \texttt{eng.gum} is English-only. Training on multiple multi-domain treebanks, including five English treebanks, did not lead to a substantial improvement in out-of-domain performance over the GUM-specific model. These findings highlight the importance of treebank-specific annotation schemes and show that the universal model remains most effective within the domains and genres present in its training data.
\begin{table}[ht]
  \centering
  \small
    \begin{tabular}{lccccc}
\toprule
 \textbf{Model} & \textbf{Seg}  & \textbf{S}  & \textbf{N}  & \textbf{R}  & \textbf{Full}  \\
\midrule
eng.gum     & \textbf{93.0}  & \textbf{58.0}  & \textbf{47.2}  & 39.1  & 38.6  \\
rus.rrg     & 85.2         & 44.7         & 34.9         & 28.8  & 28.3  \\
zho.gcdt    & 76.4         & 34.1         & 23.5         & 18.4  & 18.0  \\
UniRST      & 92.7         & 57.4         & 46.0         & \textbf{39.9}  & \textbf{39.4}   \\
\bottomrule
\end{tabular}
  \caption{Performance of the GUM-compatible models on GENTLE out-of-domain benchmark.}
  \label{tab:gentle}
\end{table}
\section{Conclusion}
While previous approaches to multilingual parsing have often advocated for reducing relation inventories to a small standardized set of RST relations, such strategies fail to fully account for the broader divergences among RST treebanks. These include differences in discourse segmentation, the treatment of mono- versus multinuclearity, and the granularity, specificity, and definitions of rhetorical relations. In this work, we introduced UniRST, the first unified RST-style discourse parser capable of effectively processing 18 treebanks across 11 languages without altering their original relation inventories. To address the challenge of inventory incompatibility, we proposed two approaches: Multi-Head and Masked-Union. Our results show that the latter yields superior performance, particularly when paired with treebank-specific segmentation heads. UniRST outperforms 16 out of 18 mono-treebank baselines, demonstrating that end-to-end multilingual discourse parsing is achievable despite considerable annotation diversity. The results indicate that embracing annotation heterogeneity can benefit multilingual discourse parsing.
\section*{Limitations}
The main limitation of a multilingual RST parser that preserves multiple relation inventories lies in the need to account for inventory differences in downstream applications. This issue is not unique to our approach, as it also arises when deploying separate treebank-specific models per language or domain. Even under label harmonization to a reduced set, variation in the number and distribution of relations across languages can persist. While UniRST demonstrates strong generalization across most treebanks, it shows a marginal performance drop on two large, multi-domain corpora (\texttt{eng.gum}, \texttt{rus.rrg}), likely because their annotations are sufficient to support strong mono-treebank models. Furthermore, \texttt{eng.sts} remains the only dataset where Span F1 remains below 50\%, with both mono- and multi-treebank models performing poorly. These observations suggest that data quality and annotation consistency substantially affect performance, and that future work may benefit from treebank filtering or weighting.
\bibliography{anthology,custom}
\appendix
\section{Reference Results from Prior Work}
\label{sec:reference-results}
Table \ref{tab:sota-compare} summarizes previously reported results for end-to-end RST parsing. It is important to note that prior results may differ in experimental setup,\footnote{Most notably, in the use of multicorpus training with harmonized label sets, or non-standard train/dev/test splits.} limiting direct comparability. All results reported in Section~\ref{ssec:monotreebank} are obtained through single-treebank training using the original relation sets and the standardized DISRPT 2025 splits. To the best of our knowledge, the remaining treebanks are evaluated here for the first time in a full end-to-end RST parsing setting.
\begin{table}[ht!]
\centering
\small
\begin{tabular}{lccccc}
\toprule
\textbf{System} & \textbf{Seg} & \textbf{S} & \textbf{N} & \textbf{R} & \textbf{Full} \\
\midrule
\multicolumn{6}{l}{\textit{eng.rstdt}} \\
SegBot \citeyearpar{zhang-etal-2020-top} & 92.2 & 62.3 & 50.1 & 40.7 & 39.6 \\
Nguyen et al. \citeyearpar{nguyen-etal-2021-rst} & 96.3 & 68.4 & 59.1 & 47.8 & 46.6 \\
DMRST \citeyearpar{liu-etal-2021-dmrst} & 96.3 & 68.4 & 59.1 & 47.8 & 46.6 \\
DMRST+ \citeyearpar{chistova-2024-bilingual} & 97.8 & 74.8 & 64.5 & 54.5 & 53.0 \\ \addlinespace
\multicolumn{6}{l}{\textit{deu.pcc}} \\
DMRST \citeyearpar{liu-etal-2021-dmrst} & 96.5 & 70.4 & 60.6 & n/c & n/c \\
\multicolumn{6}{l}{\textit{eus.ert}} \\
DMRST \citeyearpar{liu-etal-2021-dmrst} & 88.7 & 53.3 & 39.1 & n/c & n/c \\
\multicolumn{6}{l}{\textit{nld.nldt}} \\
DMRST \citeyearpar{liu-etal-2021-dmrst} & 95.5 & 62.3 & 46.6 & n/c & n/c \\
\multicolumn{6}{l}{\textit{por.cstn}} \\
DMRST \citeyearpar{liu-etal-2021-dmrst} & 92.8 & 62.5 & 51.6 & n/c & n/c \\
\multicolumn{6}{l}{\textit{rus.rrg}} \\
DMRST+ \citeyearpar{chistova-2024-bilingual} & 96.9 & 66.5 & 53.3 & 45.8 & 44.6 \\
\multicolumn{6}{l}{\textit{rus.rrt}} \\
DMRST+ \citeyearpar{chistova-2024-bilingual} & 92.2 & 65.9 & 51.0 & 43.9 & 43.8 \\
\multicolumn{6}{l}{\textit{spa.rststb}} \\
DMRST \citeyearpar{liu-etal-2021-dmrst} & 92.8 & 62.5 & 51.6 & n/c & n/c \\
\bottomrule
\end{tabular}
\caption{Reference end-to-end parsing evaluations across RST treebanks. \textbf{n/c} indicates incompatible (harmonized) label sets.}
\label{tab:sota-compare}
\end{table}
\section{Relation Classes across Treebanks}
\label{sec:appendix_labels}
\begin{figure*}[t!]
  \includegraphics[width=\linewidth]{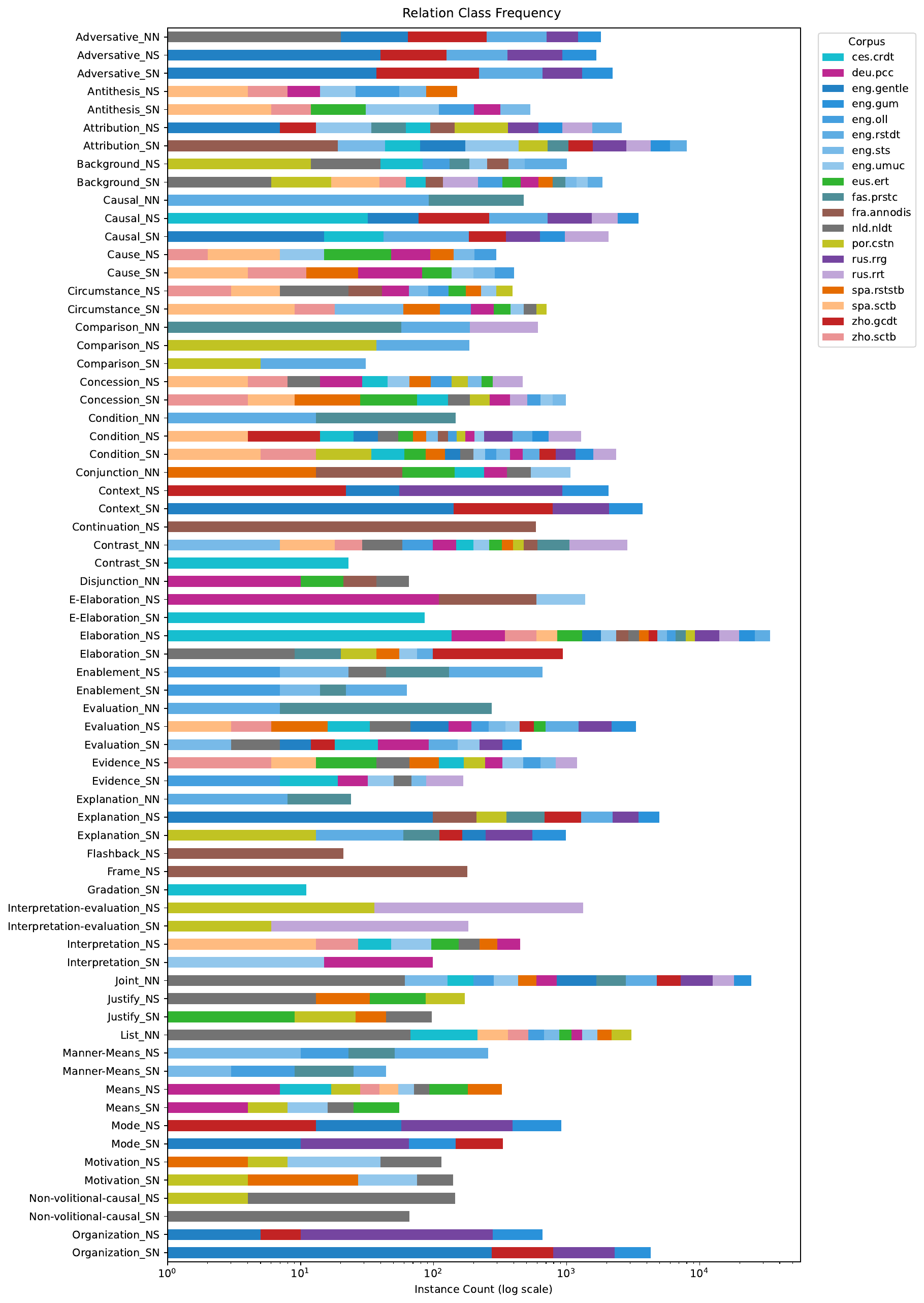}
  \caption{Relation class frequency across treebanks.}
  \label{fig:relation_distribution_part1}
\end{figure*}

\begin{figure*}[t!]
  \includegraphics[width=\linewidth]{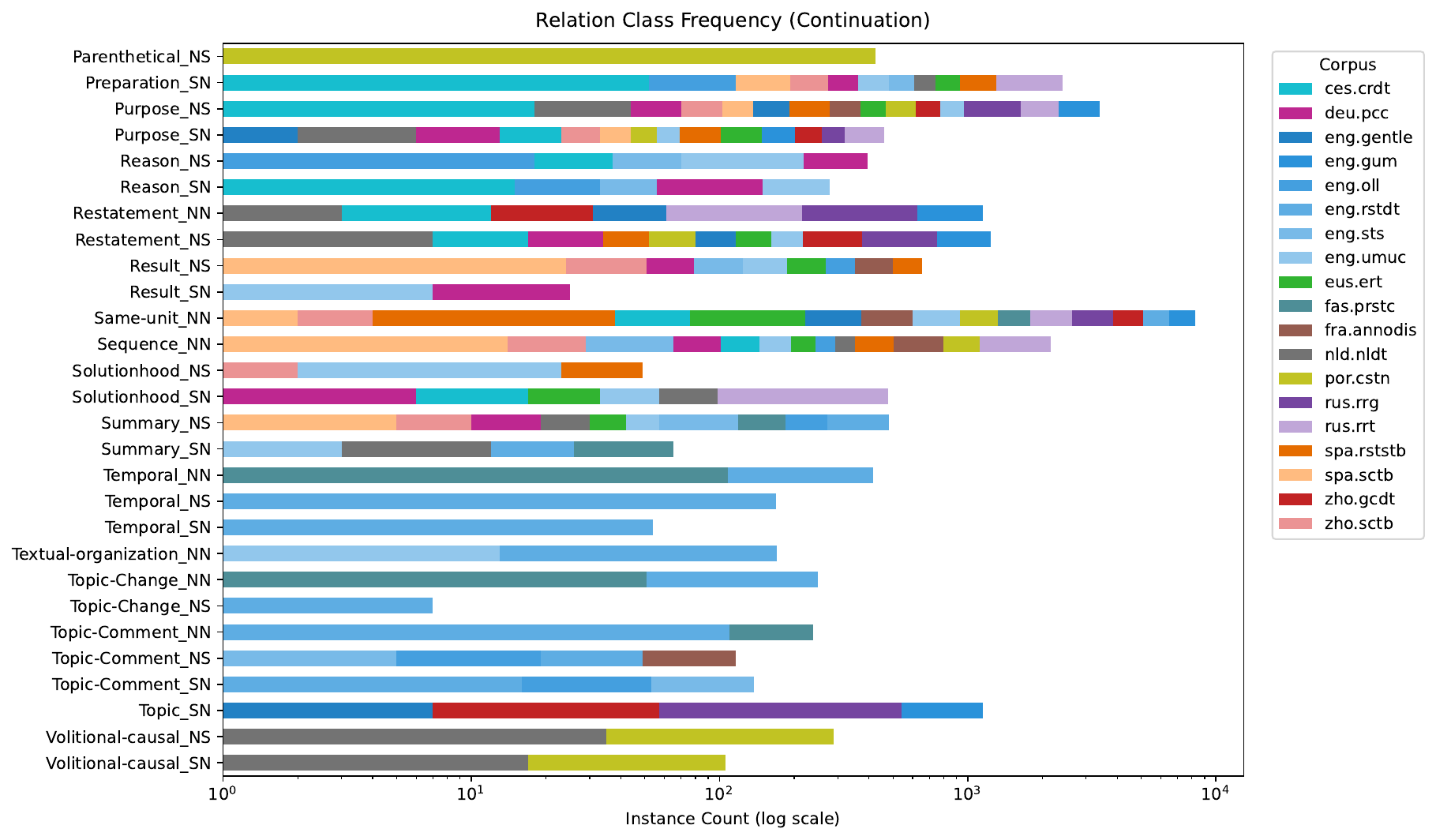}
  \caption{Relation class frequency (continuation).}
  \label{fig:relation_distribution_part2}
\end{figure*}

Figures~\ref{fig:relation_distribution_part1} and \ref{fig:relation_distribution_part2} illustrate the distribution of all relation labels across 19 treebanks (including the test‐only \texttt{eng.gentle}). UniRST handles all 96 unique \textsc{Label\_Nuclearity} relations as they appear in each corpus. Note that while some treebanks (e.g., GUM‐style and RST‑DT) internally group \textsc{Antithesis}, \textsc{Contrast}, and \textsc{Concession} as \textsc{Adversative}, and \textsc{Cause} with \textsc{Result} as \textsc{Causal}, others treat some of these relations separately or organize them under alternative groupings.
During preprocessing, only relations with equivalent definitions and comparable granularity were unified under a single label (e.g., \textsc{Condition} and \textsc{Contingency}; \textsc{Adversative} and coarse-grained \textsc{Contrast}). \textsc{Condition} is a coarse-grained label encompassing, in most treebanks, the underrepresented fine-grained relations \textsc{Otherwise}, \textsc{Unless}, and \textsc{Unconditional}, each of which appears too infrequently to be modeled reliably on its own. Labels without clear counterparts, such as \textsc{Gradation\_SN} in \texttt{ces.crdt} \cite{polakova-etal-2024-developing} or \textsc{Frame\_NS} in \texttt{fra.annodis} \cite{muller2012manuel}, remain unique to their respective treebanks. 
Variations in the representation of overlapping labels across treebanks reflect underlying genre and linguistic differences. For instance, \texttt{zho.gcdt} features more instances of \textsc{Elaboration\_SN} than \textsc{Elaboration\_NS}, in stark contrast to other languages, where the satellite in \textsc{Elaboration} typically follows the nucleus.
\end{document}